\documentclass[11pt]{article}
\makeatletter
\def\input@path{{latex/}{./}}
\makeatother

\usepackage[final]{acl}
\usepackage{dblfloatfix}
\usepackage{times}
\usepackage{latexsym}
\usepackage{amsmath}
\usepackage{algorithm}
\usepackage{algorithmic}
\usepackage{adjustbox}
\usepackage{float}
\usepackage[section]{placeins}
\usepackage{afterpage}
\usepackage{booktabs}
\usepackage[table]{xcolor}
\usepackage{listings}
\usepackage{amssymb}
\usepackage{tabularx}
\usepackage[most]{tcolorbox}
\definecolor{codekeyword}{RGB}{0, 0, 255}
\definecolor{codestring}{RGB}{163, 21, 21}
\definecolor{codecomment}{RGB}{0, 128, 0}
\definecolor{codebackground}{RGB}{255, 255, 255}
\definecolor{tblhead}{RGB}{245,245,245}
\definecolor{tblsubhead}{RGB}{250,250,250}
\definecolor{tblours}{RGB}{235,243,255}
\newcommand{\reduce}[1]{\textcolor{green!60!black}{\scriptsize\bfseries\ensuremath{\downarrow}#1}}

\newcommand\blfootnote[1]{%
  \begingroup
  \renewcommand\thefootnote{}\footnote{#1}%
  \addtocounter{footnote}{-1}%
  \endgroup
}

\lstdefinestyle{paperlisting}{
  backgroundcolor=\color{black!3},
  basicstyle=\ttfamily\footnotesize,
  numbers=none,
  frame=single,
  rulecolor=\color{black!15},
  framerule=0.4pt,
  xleftmargin=0.8em,
  xrightmargin=0.8em,
  aboveskip=0.8em,
  belowskip=0.8em,
  breaklines=true,
  breakatwhitespace=true,
  showstringspaces=false,
  tabsize=2,
  captionpos=b
}
\usepackage[T1]{fontenc}
\usepackage[utf8]{inputenc}

\usepackage{microtype}

\usepackage{inconsolata}

\usepackage{graphicx}
\graphicspath{{latex/}{./}}

\title{AgentSlimming: Towards Efficient and Cost-Aware Multi-Agent Systems}

\author{
    \textbf{Yulang Chen\textsuperscript{1,2$*$}}\quad
    \textbf{Haoxuan Peng\textsuperscript{3$*$}}\quad
    \textbf{Jinyan Liu\textsuperscript{4$*$}}\quad
    \textbf{Zichen Wen\textsuperscript{1}}\quad
\\
    \textbf{Dongrui Liu\textsuperscript{5}}\quad
    \textbf{Linfeng Zhang\textsuperscript{1$\dagger$}}
\\
\\
    \textsuperscript{1}Shanghai Jiao Tong University\quad
    \textsuperscript{2}Nanjing University\quad
    \textsuperscript{3}The University of Sydney
\\
    \textsuperscript{4}Nanjing University of Aeronautics and Astronautics
\\
    \textsuperscript{5}Shanghai AI Laboratory
\\
}

\begin{document}
\maketitle
\blfootnote{
    \begin{tabular}{@{}ll}
    \textsuperscript{$*$}Equal contribution.\\
    \textsuperscript{$\dagger$}Corresponding author (\href{mailto:zhanglinfeng@sjtu.edu.cn}{zhanglinfeng@sjtu.edu.cn}).
    \end{tabular}
}
\begin{abstract}
Large Language Model-based Multi-Agent Systems (MAS) have demonstrated remarkable capabilities in complex tasks. However, manually designing optimal communication topologies is labor-intensive, while automated expansion methods often result in bloated structures with redundant agents, leading to excessive token consumption.
To address this problem, we introduce \textbf{AgentSlimming}, a plug-and-play compression framework for graph-structured multi-agent workflows. Motivated by pruning and quantization in neural networks, AgentSlimming compresses workflows by first estimating the importance score of each agent with a hybrid mechanism, and then removes redundant agents or replaces them with low-cost ones, where each operation is validated using a baseline-anchored acceptance rule to prevent performance collapse. Experiments show that AgentSlimming reduces average token cost by up to 78.9\% with negligible performance degradation, and sometimes even improves accuracy, achieving a strong Pareto-optimal trade-off between cost and quality. \textit{Our code is publicly available at \url{https://github.com/CitrusYL/AgentSlimming}.}
\end{abstract}

\section{Introduction}

% These instructions are for authors who submit papers to *ACL conferences using \LaTeX. They are not self-contained. All authors must follow the general instructions for *ACL proceedings \url{http://acl-org.github.io/ACLPUB/formatting.html},and this document contains additional instructions for the \LaTeX{} style files.

% The templates include the \LaTeX{} source of this document (\texttt{acl\_latex.tex}),
% the file of \LaTeX{} style used to format it (\texttt{acl.sty}),
% an ACL bibliography style (\texttt{acl\_natbib.bst}),
% an example bibliography (\texttt{custom.bib}),
% and the bibliography for the ACL Anthology (\texttt{anthology.bib}).

The paradigm of Multi-Agent Systems (MAS) has shifted from manually crafting static prompts to orchestrating dynamic collaborations among specialized agents. Frameworks such as AutoGen \citep{wu2023autogen} and ADAS \citep{hu2024adas} have shown that decomposing complex problems into sub-tasks can improve performance. Recent innovations, such as MetaGPT \citep{hong2024metagpt} and AFlow \citep{zhang2025aflow}, have revolutionized this field by reformulating workflow generation as a search problem. By leveraging Monte Carlo Tree Search (MCTS), AFlow can automatically navigate the vast space of agent interactions to discover highly effective reasoning topologies.

\begin{figure}[t]
  \centering
  \vspace{-0.1cm}\includegraphics[width=\columnwidth]{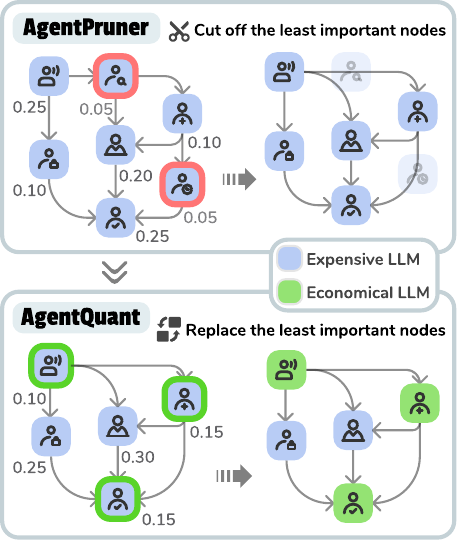}
  \caption{\textbf{Overview of the pruning and quantization concepts in AgentSlimming.} The process begins with a workflow that is initially high-performance but computationally expensive. Then, a \emph{hybrid importance evaluation mechanism} is utilized to calculate each agent node's importance score (\emph{i.e.,} the float value), which guides \textit{pruning} and \textit{quantization} in AgentSlimming.}
  \label{fig:pruning&quantization}
  \vspace{-0.6cm}
\end{figure}

However, this automated discovery process operates under an "unconstrained resources" assumption. The resulting workflows, while accurate, tend to be computationally bloated. They often feature dense, redundant connectivity and uniformly employ the most capable and expensive LLMs, e.g., GPT-4 \citep{openai2023gpt4}, as agent nodes for every sub-task \citep{li2023camel, wei2024chatdev}. This leads to two critical bottlenecks: \textbf{redundant communication}, where agents exchange repetitive or low-value information, and \textbf{excessive computational cost}, where simple sub-tasks consume high-value resources. As the number of agents and interaction turns grows, a quadratic increase in token consumption renders these systems difficult to scale.

To solve this problem, we propose AgentSlimming, a framework that can be directly applied to any MAS and workflows to reduce their execution costs. Motivated by the concepts of pruning \citep{han2015efficient, frankle2019lottery}, quantization \citep{jacob2018quant, dettmers2022int8}, and finetuning \citep{howard2018ft} in traditional neural network compression, we define our pruning as the removal of an agent node, and semantic quantization as replacing the LLM associated with an agent node with a lower-cost model. AgentSlimming introduces AgentPruner, AgentQuant and AgentTuner as tools towards lightweight MAS.
As shown in Figure~\ref{fig:pruning&quantization}, AgentPruner identifies the least important agent nodes in the MAS and removes them, while AgentQuant replaces the underlying LLMs of the remaining lower-importance nodes with more cost-effective alternatives. When high-cost, high-precision models (e.g., GPT-4) are considered unnecessary for a given role in MAS, they are dynamically replaced by cost-effective, lightweight surrogates (e.g., GPT-4o-mini) to reduce the inference costs. 
After AgentPruner and AgentQuant, AgentTuner is optionally employed to recover performance loss and further optimize.

The proposition of AgentSlimming introduces a crucial problem for MAS compression, \emph{i.e.,} how to determine the importance of each node in MAS. To tackle this challenge, we introduce a hybrid importance evaluation mechanism. Generally, a node's value is determined by both its \emph{structural position} and \emph{functional contribution}. Based on this observation, we consider the following three indicators: Degree Centrality \citep{freeman1978centrality} and Betweenness Centrality \citep{brandes2001faster} to capture the topological structure in the graph of MAS, and an Approximate Shapley value \citep{shapley1953value, lundberg2017unified} for functional contribution. Specifically, we adopt a Leave-One-Out (LOO) estimation strategy to calculate the marginal contribution of each node against the complete topology, serving as a computationally efficient proxy for the exact Shapley value. These rankings are fused via Reciprocal Rank Fusion (RRF) \citep{cormack2009reciprocal} to generate a robust importance score. 

Guided by this score, AgentSlimming employs an iterative greedy strategy: we progressively prune and quantize agent nodes starting from the least important ones. After each operation, we re-evaluate the workflow against a performance threshold and re-calculate the RRF scores, ensuring the optimization dynamically adapts to the changing graph structure until the efficiency limit is reached.
While the pruned and quantized workflows typically achieve a superior Pareto frontier \citep{deb2002fast, chen2024frugalgpt} between cost and accuracy, AgentSlimming further incorporates a MCTS fine-tuning mechanism that optimizes the streamlined graph rather than just accepting the compressed policy. Empirically, the resulting workflows are consistently more cost-efficient than AFlow, often achieving strictly superior performance in both accuracy and cost. In terms of average execution cost per problem (USD), AgentSlimming demonstrates significant gains across diverse benchmarks.

For instance, on GSM8K, AgentSlimming matches AFlow's score (95.5) while reducing the cost by \textbf{78.8\%}. On code generation and complex reasoning tasks like MBPP and LiveCode, AgentSlimming achieves a "dual victory": it improves accuracy while reducing costs by \textbf{71.7\%} and \textbf{78.9\%}, respectively. Even in cases where AFlow retains a marginal score advantage (e.g., HotpotQA: 77.3 vs.\ 77.0), AgentSlimming achieves this at a substantially lower cost (27.4\% reduction), indicating that our pipeline reliably locates the workflow at a more favorable operating point on the cost--quality Pareto frontier. Our contributions can be summarized as follows:
\begin{itemize}
    \item We introduce AgentSlimming, a training-free framework that integrates automated agentic workflow exploration with a novel pruning and semantic quantization pipeline.
    \item We propose a novel iterative multi-metric optimization algorithm that integrates topological and functional importance via RRF to accurately identify redundant components.
    \item Extensive evaluations on eight benchmarks show that AgentSlimming achieves a striking reduction in token costs of up to \textbf{78.9\%}.
\end{itemize}

\section{Related Work}

\subsection{Agentic Workflows }
LLM-based systems can be broadly characterized into two paradigms: \emph{agentic workflows} and \emph{autonomous agents}. The former executes tasks through predefined multi-step pipelines with repeated LLM invocations, whereas the latter emphasizes dynamic decision-making and planning through interaction and feedback. Recent work has made notable progress in language-driven decomposition and collaboration \citep{zhuge2023mindstorms}, data-science agents \citep{hong2024datainterpreter}, tool-enabled mobile agent teams \citep{zhang2024mobileexperts}, and open-ended embodied exploration \citep{wang2023voyager}. Compared to autonomous agents that often require environment-specific action spaces and decision patterns, workflows can more easily incorporate human domain expertise and improve through iterative refinement, making them particularly amenable to automated construction and optimization.

\subsection{Multi-Agent Evolving}
Many effective workflows are still developed primarily through manual discovery and engineering practice. At the general level, common transferable reasoning recipes include chain-of-thought, self-consistency, self-refine and structured self-collaboration \citep{wei2022cot, wang2022selfconsistency, amen2023selfrefine, wang2024cognitivesynergy}. At the domain level, multi-step procedures are organized into reusable pipelines, such as code generation and debugging \citep{hong2024metagpt, ridnik2024alphacodium, zhong2024ldb}, data analysis and visualization \citep{xie2024haichart, ye2024genaiviz, li2024nl2sql, zhou2023llmdba}, mathematical reasoning \citep{xu2024lemur}, and planning-style search for problem solving \citep{nori2023medprompt, zhou2024lats}.

However, manual design cannot cover the combinatorial space across domains, which motivates automated agentic optimization. One line of work optimizes local instructions or components within a fixed backbone \citep{fernando2024promptbreeder, yuksekgonul2024textgrad, yang2024llmasoptimizers, khattab2024dspy} or tunes inference-time strategies \citep{saadfalcon2024archon}. Another line goes further by optimizing the end-to-end workflow structure, including automatic generation of code-represented workflows \citep{li2024autoflow}, viewing agent systems as optimizable graphs \citep{zhuge2024gptswarm}, and improving system designs in code space via meta-agents \citep{hu2024adas}. \citet{zhang2025aflow} similarly adopts code-based representations, combining finer-grained abstractions (named nodes/operators) with MCTS to leverage execution feedback and tree-structured experience for efficient workflow structure search.

\subsection{Graph-Structured Orchestration, Pruning, and Cost-Aware Compression}
Beyond prompt engineering, many agentic systems are best viewed as executable graphs, where nodes represent specialized modules and edges encode information flow. Recent studies focus on topology learning and pruning for improved efficiency and robustness \citep{zhang2024cutthecrap, zhang2024gdesigner, li2025agp}. These approaches also explore pruning at different granularities, including message-level pruning under bandwidth constraints \citep{mao2020messagepruning}, dynamic agent elimination for token efficiency \citep{wang2025agentdropout}, and progressive pruning that blends heuristics with execution experience \citep{zhang2025safesieve}. Existing methods typically start from (near) fully connected interactions and primarily prune edges or communication channels. Action selection for pruning and replacement can draw on heuristic importance measures from network analysis \citep{freeman1978centrality} or contribution-based formulations such as Shapley values \citep{shapley1953value}, which usually require approximation for tractability. In addition, model compression and quantization reduce inference cost while preserving quality \citep{frantar2022gptq, xiao2023smoothquant, lin2023awq}, complementing structural optimization at the workflow level.

\subsection{Search-Based Optimization for Workflows}
Search provides a natural mechanism for exploring large workflow design spaces. MCTS \citep{coulom2006mcts} and its variant UCT \citep{kocsis2006uct} enable effective exploration through sampled evaluation and progressive expansion, and has been applied to planning and reasoning in language agents \citep{zhou2024lats} as well as workflow structure optimization \citep{zhang2025aflow}. Following this trajectory, we focus on optimizing DAG workflows under dependency constraints, integrating structure search with node-level pruning and cost-driven node replacement (semantic quantization).

\section{Methodology}

\begin{figure*}[t]
  \centering
  \includegraphics[width=\textwidth]{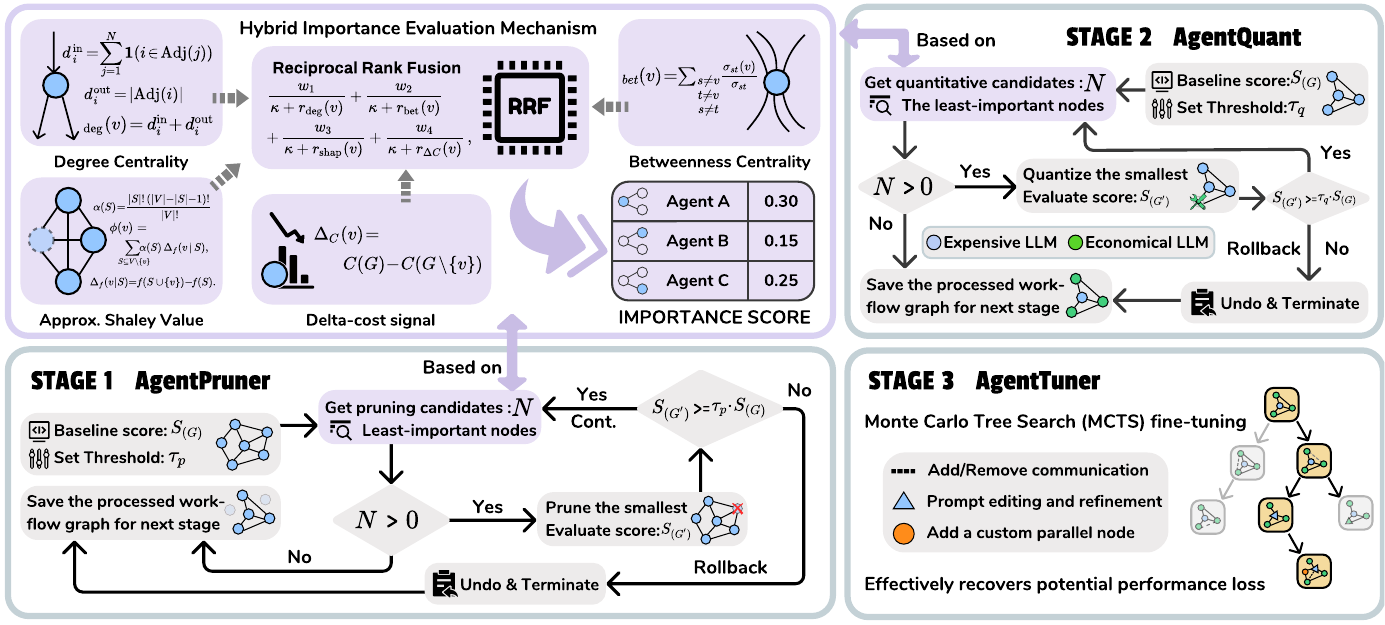}
  \caption{\textbf{Illustration of AgentSlimming.} To identify redundancy, AgentSlimming computes four distinct rankings of each agent node: \textit{Degree Centrality}, \textit{Betweenness Centrality}, \textit{Cost Comparison} and \textit{Approximate Shapley value}. Both pruning and quantization select candidate nodes based on the computed importance scores, ranking nodes from least to most important and optimizing the lowest-ranked candidate first. After each operation, a re-evaluation is performed. If the score drops below an acceptable threshold, a rollback mechanism is triggered to revoke the current operation and immediately terminate this phase, thereby preserving superior performance.}
  \label{fig:fig2_pipeline}
\end{figure*}

\subsection{Problem Formulation}

We formulate the optimization of multi-agent systems as a discrete structural search problem over a directed graph, with an explicit trade-off between task performance and execution cost.
\paragraph{Workflow as a directed graph.}
A multi-agent workflow is represented as a directed graph $\mathcal{G}=(\mathcal{V}, \mathcal{E})$. Each node $v\in \mathcal{V}$ denotes an executable operator (e.g., an LLM call), and each edge $e\in \mathcal{E}$ specifies an information dependency. Given a validation set $\mathcal{D}_{\mathrm{val}}$, executing $\mathcal{G}$ yields an average task score $S(\mathcal{G})$ and an average execution cost $C(\mathcal{G})$:
\begin{equation}
\begin{split}
S(\mathcal{G}) &= \frac{1}{\lvert \mathcal{D}_{\mathrm{val}}\rvert}\sum_{x\in \mathcal{D}_{\mathrm{val}}}\mathrm{score}(\mathcal{G},x),\\
C(\mathcal{G}) &= \frac{1}{\lvert \mathcal{D}_{\mathrm{val}}\rvert}\sum_{x\in \mathcal{D}_{\mathrm{val}}}\mathrm{cost}(\mathcal{G},x).
\end{split}
\end{equation}

\paragraph{Optimization Objective.}
Our goal is to maximize performance under a cost budget $B$. When multiple topologies achieve comparable performance, we prioritize the one with the minimum cost:
\begin{equation}
\max_{\mathcal{G}}\; S(\mathcal{G}) \quad \text{s.t.} \quad C(\mathcal{G}) \le B.
\end{equation}

\subsection{The AgentSlimming Pipeline}

As illustrated in Figure~\ref{fig:fig2_pipeline}, our framework operates through a three-stage pipeline: \textit{AgentPruner (Structural Pruning)}, \textit{AgentQuant (Semantic Quantization)}, and \textit{AgentTuner (MCTS Fine-tuning)}. This pipeline progressively compresses the workflow while maintaining its reasoning capabilities.

\paragraph{Hybrid Importance Evaluation} To guide both pruning and quantization, we introduce a hybrid importance evaluation mechanism. For any node \(v \in \mathcal{V}\), we compute four complementary signals using a small probe dataset \(\mathcal{D}_{\mathrm{probe}} \subset \mathcal{D}_{\mathrm{val}}\). To evaluate the sensitivity of our method to the probe dataset size, we further conduct additional experiments on HotpotQA, with details provided in Appendix~\ref{app:probe_size}.

\paragraph{1. Degree and Betweenness Centrality Signals (Topological Priors)}
We capture the structural significance of a node using graph centrality metrics.
\begin{itemize}
    \item \textbf{Degree Centrality Signal ($s_{\mathrm{deg}}$):} It is derived from the node's local connectivity (in-degree and out-degree). Nodes with weaker structural connectivity are assigned larger pruning likelihood.
    \item \textbf{Betweenness Centrality Signal ($s_{\mathrm{bet}}$):} Measures the node's role as an information bridge. It is calculated as:
    \begin{equation}
    s_{\mathrm{bet}}(v) = \sum_{s \neq v \neq t} \frac{\sigma_{st}(v)}{\sigma_{st}},
    \end{equation}
    where $\sigma_{st}$ is the total number of shortest paths from $s$ to $t$, and $\sigma_{st}(v)$ is the number of those paths passing through $v$.
\end{itemize}

\paragraph{2. Approximate Shapley value Signal (Functional Contribution)}
To quantify the semantic contribution of node $v$, we adopt the Shapley value concept from cooperative game theory. Since exact computation is NP-hard, we employ an efficient \textbf{Leave-One-Out (LOO)} approximation. The marginal contribution $\hat{\phi}(v)$ is estimated by the performance drop when $v$ is removed (or replaced):
\begin{equation}
\hat{\phi}(v) \approx \Delta_S(v) = S(\mathcal{G}) - S(\mathcal{G} \setminus \{v\}),
\end{equation}
where $\mathcal{G} \setminus \{v\}$ denotes the workflow after removing node $v$ (and patching connections). A smaller $\hat{\phi}(v)$ indicates a lower contribution to task completion.

\paragraph{3. Delta-Cost Signal (Economic Potential)}
We estimate the potential cost saving $\Delta_C(v)$ on $\mathcal{D}_{\mathrm{probe}}$:
\begin{equation}
\Delta_C(v) = C(\mathcal{G}) - C(\mathcal{G} \setminus \{v\}),
\end{equation}
where $C(\cdot)$ denotes the monetary API cost, measured in USD per problem, incurred when executing the workflow on $\mathcal{D}{\mathrm{probe}}$. Specifically, it is computed based on the actual API billing rate and the total token usage (input and output tokens) of all model calls within the workflow. Nodes with higher $\Delta_C(v)$ are prioritized for pruning or quantization to maximize efficiency gains.

\paragraph{Rank Fusion}
To robustly combine these heterogeneous signals, we employ Reciprocal Rank Fusion (RRF). Let $r_m(v)$ be the rank of node $v$ under metric $m \in \{\mathrm{deg}, \mathrm{bet}, \mathrm{shap}, \Delta C\}$. The fused score is calculated as:
\begin{equation}
\mathrm{RRF}(v) = \sum_{m} \frac{w_m}{\kappa + r_m(v)},
\end{equation}
where $\kappa$ is a smoothing constant (set to $\max\{10, |\mathcal{V}|\}$) and $w$ controls the relative weight of each metric. This fused score identifies nodes that are simultaneously structurally peripheral, functionally redundant, and computationally expensive.

\subsubsection{Iterative Optimization Process}

\paragraph{Stage 1: AgentPruner (Structural Pruning)} Let $\mathcal{G}_{\mathrm{base}}$ denote the initial high-performance workflow graph. We employ an iterative greedy strategy to prune nodes. In each iteration, we rank nodes by $\mathrm{RRF}(v)$ and evaluate the top candidates. A pruning operation is accepted if the performance degradation is within a tolerance threshold $\tau_p$:
\begin{equation}
S(\mathcal{G}_{p}) \ge \tau_p \cdot S(\mathcal{G}_{\mathrm{base}}),
\end{equation}
where $\mathcal{G}_{p}$ is the pruned graph and $\tau_p\in(0,1)$ is a predefined threshold. We evaluate the Top-1 candidate first and then the remaining candidates in order if necessary. If neither candidate is accepted, the stage terminates and the workflow rolls back to the last accepted state. \textbf{Graph Surgery:} To preserve executability after removing node $v$, we perform edge patching to maintain the graph's connectivity. For every pair of predecessor $s \in \mathrm{In}(v)$ and successor $t \in \mathrm{Out}(v)$, we add a direct edge $(s \to t)$ if one does not already exist, where self-loops and duplicate edges are disallowed. Each accepted pruning step is logged with the probe signals, fused ranks, and the resulting workflow artifact.

\paragraph{Stage 2: AgentQuant (Semantic Quantization)} $\mathcal{G}_{p}$ refers to the pruned workflow graph obtained from Stage 1. The quantization process operates on this sparse structure to further reduce computational costs. We define \textit{Semantic Quantization} as a model substitution strategy. For a selected node $v$, we replace its high-cost LLM with a cost-effective surrogate $\pi(v)$, yielding a quantized graph $\mathcal{G}_q$. Candidate ranking follows the same RRF mechanism, where the Shapley signal is instantiated by evaluating the performance impact of the model substitution. We also employ a greedy strategy to quantize nodes. A quantization operation is accepted if the performance degradation is within a tolerance threshold $\tau_q$:
\begin{equation}
S(\mathcal{G}_{q}) \ge \tau_q \cdot S(\mathcal{G}_{p}),
\end{equation}
where $\mathcal{G}_{q}$ is the quantized graph and $\tau_q\in(0,1)$ is a predefined threshold. This step drastically reduces token costs for non-critical reasoning steps. We employ a "greedy + rollback" optimization strategy, as global combinatorial search on large graphs is computationally prohibitive. Although this risks missing complex multi-node synergies, we mitigate this limitation by using approximate Shapley values and relaxed thresholds to balance efficiency with effective exploration.

\paragraph{Stage 3: AgentTuner (Adaptive MCTS Fine-tuning)} To address the variance in compression sensitivity across different tasks, we adapt the MCTS exploration from AFlow \citep{zhang2025aflow} for our compressed workflow. This phase is selectively applied and focuses on localized refinements, such as prompt adaptation and parameter tuning, to recover from potential degradation. Critically, this fine-tuned configuration is adopted only if it improves upon the initial compressed version, thereby ensuring consistent and robust performance gains.

\FloatBarrier
\section{Experiments}

\subsection{Setup}
\label{sec:setup}

\paragraph{Overview.}
We evaluate our method on eight public benchmarks. To ensure a fair comparison, we strictly adhere to AFlow's configuration, adopting its datasets, splits, and sampling methods. Additionally, we incorporate three challenging benchmarks to test robustness.

\begin{table*}[t]
  \centering
  \small
  \setlength{\tabcolsep}{2.5pt}        % column padding
  \renewcommand{\arraystretch}{1.15}   % row padding

  \begin{tabular}{lcccccccccc}
    \toprule
    \rowcolor{tblhead} 
    \textbf{Method}
      & \multicolumn{2}{c}{\textbf{MATH}}
      & \multicolumn{2}{c}{\textbf{GSM8K}}
      & \multicolumn{2}{c}{\textbf{HotpotQA}}
      & \multicolumn{2}{c}{\textbf{MBPP}}
      & \multicolumn{2}{c}{\textbf{DROP}} \\
    \cmidrule(lr){2-3}\cmidrule(lr){4-5}\cmidrule(lr){6-7}\cmidrule(lr){8-9}\cmidrule(lr){10-11}

    \rowcolor{tblsubhead}
      & Acc. & Cost (\$/prob.)
      & Acc. & Cost (\$/prob.)
      & Acc. & Cost (\$/prob.)
      & Acc. & Cost (\$/prob.)
      & Acc. & Cost (\$/prob.) \\
    \midrule

    \rowcolor{tblours}
    \textbf{Ours}
      & 73.9 & 6.88\text{e-}3 \reduce{42.7\%}
      & \textbf{95.5} & 9.30\text{e-}4 \reduce{78.8\%}
      & 77.0 & 5.55\text{e-}3 \reduce{27.4\%}
      & \textbf{77.9} & 7.30\text{e-}4 \reduce{71.7\%}
      & \textbf{81.7} & 9.50\text{e-}4 \reduce{38.7\%} \\
    
    AFlow
      & \textbf{74.8} & 1.20\text{e-}2
      & \textbf{95.5} & 4.38\text{e-}3
      & \textbf{77.3} & 7.64\text{e-}3
      & 73.3 & 2.58\text{e-}3
      & 78.7 & 1.55\text{e-}3 \\
    LLM-Debate
      & 74.5 & 1.47\text{e-}2
      & 93.5 & 3.57\text{e-}3
      & 73.4 & 9.47\text{e-}3
      & 75.1 & 6.22\text{e-}3
      & 75.6 & 2.94\text{e-}3\\
    ADAS
      & 69.0 & 1.15\text{e-}2
      & 94.9 & 4.79\text{e-}3
      & 65.4 & 4.72\text{e-}2
      & 75.4 & 4.31\text{e-}3
      & 78.6 & 3.07\text{e-}3 \\
    
    CoT
      & 62.9 & 6.03\text{e-}4
      & 90.5 & 2.48\text{e-}4
      & 57.5 & 7.11\text{e-}4
      & 73.6 & 2.80\text{e-}4
      & 75.9 & 3.75\text{e-}4 \\
    
    SC-CoT
      & 65.3 & 3.67\text{e-}3
      & 90.2 & 1.71\text{e-}3
      & 61.7 & 4.48\text{e-}3
      & 75.6 & 2.12\text{e-}3
      & 75.3 & 2.37\text{e-}3 \\
    
    Self-Refine
      & 65.5 & 1.01\text{e-}3
      & 89.8 & 5.60\text{e-}4
      & 57.9 & 1.50\text{e-}3
      & 74.5 & 5.56\text{e-}4
      & 75.0 & 7.91\text{e-}4 \\
    
    \bottomrule
  \end{tabular}

  \caption{\textbf{Performance and inference cost comparison across standard benchmarks.} Cost denotes the average API expense (\$/prob.). \textbf{Bold} indicates the best results among \textbf{workflow-based methods} (excluding simple prompting baselines). Green percentages indicate the relative cost reduction compared to AFlow, highlighting how much inference budget is saved while maintaining competitive accuracy.}
  \label{tab:main-results}
\end{table*}

\begin{table*}[t]
  \centering
  \small
  \setlength{\tabcolsep}{4pt}          % column padding
  \renewcommand{\arraystretch}{1.15}   % row padding

  \begin{tabular}{lcccccc}
    \toprule
    \rowcolor{tblhead}
    \textbf{Method}
    & \multicolumn{2}{c}{\textbf{AIME}}
    & \multicolumn{2}{c}{\textbf{LiveCode}}
    & \multicolumn{2}{c}{\textbf{MusiqueAns}} \\
    
    \cmidrule(lr){2-3} \cmidrule(lr){4-5} \cmidrule(lr){6-7}
    
    \rowcolor{tblsubhead}
    & Acc. & Cost (\$/prob.)
    & Acc. & Cost (\$/prob.)
    & Acc. & Cost (\$/prob.) \\
    
    \midrule

    \rowcolor{tblours}
    \textbf{Ours}
    & 65.7 & 1.35\text{e-}2 \reduce{19.2\%}
    & 61.7 & 2.47\text{e-}3 \reduce{78.9\%}
    & \textbf{89.3} & 8.24\text{e-}3 \reduce{23.0\%} \\
    
    AFlow
    & \textbf{67.1} & 1.67\text{e-}2
    & 55.3 & 1.17\text{e-}2
    & 84.5 & 1.07\text{e-}2 \\
    
    LLM-Debate
      & 56.8 &  2.86\text{e-}2
      & \textbf{67.8} & 1.67\text{e-}2
      & 45.3 &  1.62\text{e-}2\\
    Cost-aware AFlow
    & 61.8 & 1.20\text{e-}2
    & 57.9 & 8.41\text{e-}3
    & 83.6 & 1.04\text{e-}2  \\
    
    ADAS
    & 62.3 & 1.42\text{e-}2
    & 52.1 & 9.01\text{e-}3
    & 77.2 & 1.62\text{e-}2 \\
    
    CoT
    & 53.4 & 2.99\text{e-}3
    & 46.4 & 1.25\text{e-}3
    & 73.9 & 1.11\text{e-}3 \\
    
    Self-Consistency CoT
    & 59.7 & 1.41\text{e-}2
    & 47.2 & 7.93\text{e-}3
    & 74.9 & 6.89\text{e-}3 \\
    
    Self-Refine
    & 54.1 & 4.08\text{e-}3
    & 49.1 & 2.30\text{e-}3
    & 72.8 & 2.30\text{e-}3 \\
    
    \bottomrule
  \end{tabular}
  \caption{\textbf{Performance and inference cost comparison across high-difficulty benchmarks.} Cost denotes the average API cost (\$/prob.). \textbf{Bold} indicates the best results among \textbf{workflow-based methods} (excluding simple prompting baselines). Green percentages indicate the relative cost reduction compared to AFlow.}
  \label{tab:hard-benchmarks}
\end{table*}

\paragraph{Datasets.}
Our evaluation suite consists of two categories:
(1) \textbf{Standard Benchmarks.} We utilize the complete AFlow \citep{zhang2025aflow} suite, including GSM8K \citep{zhong2021gsm8k}, MBPP \citep{austin2021mbpp} (full sets), as well as HotpotQA \citep{yang2018hotpotqa} and DROP \citep{dua2019drop} (randomly sampled 1,000-instance subsets). For MATH \citep{dan2021math}, we follow the specific subset of 617 Level-5 problems across four categories. These datasets employ a 1:4 validation/test split.
(2) \textbf{High-difficulty Benchmarks.} To assess performance on complex reasoning and coding tasks, we incorporate AIME~\cite{maa_aime,aops_aime_archive}
, MuSiQueAns \citep{harsh2022musiqueans}, and LiveCode \citep{naman2025livecode}. These datasets are partitioned using a 3:7 validation/test split. Across both categories, our evaluation suite collectively covers three core competencies: mathematical reasoning, coding, and question answering, ensuring a comprehensive assessment of agent system performance.

\paragraph{Baselines.}
We compare AgentSlimming against a diverse set of baselines:
\emph{(I) Manual Prompting Strategies:} Standard Chain-of-Thought (CoT) \citep{wei2022cot}, Self-Consistency (SC-CoT) \citep{wang2022selfconsistency}, Self-Refine \citep{amen2023selfrefine}, and LLM-Debate \citep{du2023improving}.
\emph{(II) Automated Workflow Optimization:} ADAS \citep{hu2024adas} and AFlow \citep{zhang2025aflow}.
\emph{(III) Cost-Aware AFlow:} We implement a modified AFlow variant that explicitly integrates cost comparisons into its selection mechanism during the search phase, directly competing on cost efficiency.

\paragraph{Details.}
We employ GPT-4.1-mini \citep{openai2025gpt41} as the high-precision model and GPT-4.1-nano \citep{openai2025gpt41} as the cost-effective surrogate for quantization, both accessed via the OpenAI API. For workflow-based optimization, we use GPT-4.1-mini to maintain computational efficiency. All models are accessed with temperature set to 0 to ensure reproducibility. Full hyperparameter configurations are provided in Appendix~\ref{app:details}. We further extend our study with cross-architecture experiments on the Qwen3 series, cross-dataset validation on GSM8K and MATH, and additional experiments based on ADAS, which demonstrate that AgentSlimming is not tied to AFlow-specific design choices; detailed results are provided in Appendix~\ref{app:generalization}.

\subsection{Main Results}
\paragraph{Superior Cost-Performance Efficiency.} The main experimental results are summarized in Table~\ref{tab:main-results} and Table~\ref{tab:hard-benchmarks}. Overall, AgentSlimming consistently discovers workflows that are more cost-efficient than AFlow. In many cases, our method achieves \textit{strict dominance}, surpassing the baseline in both accuracy and execution cost. Here, we report the average cost in USD per problem.

On standard benchmarks, AgentSlimming maintains competitive performance while significantly reducing computational overhead. For instance, on GSM8K, it matches AFlow’s accuracy (95.5) but reduces the average cost from 4.38$\times10^{-3}$ to 9.30$\times10^{-4}$, a \textbf{78.8\%} cost reduction. On MBPP, AgentSlimming improves the score from 73.3 to 77.9 while simultaneously lowering the cost from 2.58$\times10^{-3}$ to 7.30$\times10^{-4}$ (\textbf{71.7\%} reduction). On datasets where AFlow attains a slightly higher raw score, such as HotpotQA, AgentSlimming achieves comparable performance at a substantially lower cost, reducing the average cost by 27.4\%.

The advantage of AgentSlimming is even more pronounced on high-difficulty benchmarks. On LiveCode, it increases the score from 55.3 to 61.7 while reducing cost by nearly an order of magnitude (from 1.17$\times10^{-2}$ to 2.47$\times10^{-3}$, a \textbf{78.9\%} reduction). Similarly, on MuSiQueAns, AgentSlimming improves accuracy from 84.5 to 89.3 with a \textbf{23.2\%} reduction. These results indicate that AgentSlimming reliably positions the discovered workflows at more favorable operating points on the cost--quality Pareto frontier, particularly for complex reasoning and programming tasks.

\definecolor{tblorder}{rgb}{0.85, 0.92, 1}

\begin{table*}[t]
\centering
\footnotesize
\setlength{\tabcolsep}{6pt}
\renewcommand{\arraystretch}{1.10}

\begin{tabular}{lcccccc}
    \toprule
    \rowcolor{tblhead}
    \textbf{Method}
    & \multicolumn{2}{c}{\textbf{MATH}}
    & \multicolumn{2}{c}{\textbf{MBPP}}
    & \multicolumn{2}{c}{\textbf{LiveCode}} \\
    \cmidrule(lr){2-3} \cmidrule(lr){4-5} \cmidrule(lr){6-7}
    & Acc. & Cost (\$/prob.)
    & Acc. & Cost (\$/prob.)
    & Acc. & Cost (\$/prob.) \\
    \midrule

    Baseline
    & 74.8 & 1.20e-2
    & 73.3 & 2.58e-3
    & 55.3 & 1.17e-2 \\

    Quantization -> Pruning
    & 76.5 & 3.05e-3
    & 75.6 & 6.30e-4
    & 59.8 & 1.99e-3 \\

    \rowcolor{tblorder}
    Pruning -> Quantization (Original Method)
    & 73.9 & 6.88e-3
    & 77.9 & 7.30e-4
    & 61.7 & 2.47e-3 \\

    \bottomrule
\end{tabular}

\caption{\textbf{Ablation of reversing the pipeline order.} We compare the baseline, the reversed-order pipeline (\textit{Pruning after Quantization}), and the final \textit{Quantization after Pruning} results on MATH, MBPP, and LiveCode, which represent mathematical reasoning, program synthesis, and execution-oriented code generation, respectively. Empirically, reversing the order produces a broadly similar final cost-accuracy trade-off, with only minor differences arising from slightly different search trajectories under the greedy rollback mechanism. We nevertheless adopt the \textit{Pruning $\rightarrow$ Quantization} design in our framework, because pruning removes redundant nodes early, shrinks the search space, and thus makes the subsequent quantization stage more efficient and reduces the overall search cost of the framework.}
\label{tab:order_ablation}
\end{table*}

\definecolor{tblhead}{RGB}{245,245,245}
\definecolor{tblRRF}{rgb}{0.85, 0.92, 1}

\begin{table*}[t]
\centering
\scriptsize
\setlength{\tabcolsep}{4pt}
\renewcommand{\arraystretch}{1.08}

\begin{tabular}{@{}lcccccccc@{}}
    \toprule
    \rowcolor{tblhead}
    \textbf{Method}
    & \multicolumn{2}{c}{\textbf{AIME}}
    & \multicolumn{2}{c}{\textbf{DROP}}
    & \multicolumn{2}{c}{\textbf{MATH}}
    & \multicolumn{2}{c}{\textbf{MBPP}} \\
    \cmidrule(lr){2-3} \cmidrule(lr){4-5} \cmidrule(lr){6-7} \cmidrule(lr){8-9}
    & Acc. & Cost
    & Acc. & Cost
    & Acc. & Cost
    & Acc. & Cost \\
    \midrule
    
    Baseline
    & 67.1 & 1.67e-2
    & 78.7 & 1.55e-3
    & 74.8 & 1.20e-2
    & 73.3 & 2.58e-3 \\

    Betweenness-only
    & 63.0 & 1.15e-2
    & 77.8 & 4.5e-4
    & 74.0 & 4.17e-3
    & 79.1 & 7.1e-4 \\

    Degree-only
    & 63.6 & 7.89e-3
    & 66.5 & 3.2e-4
    & 73.1 & 5.28e-3
    & 73.3 & 1.56e-3 \\

    Shapley-only
    & 64.7 & 7.81e-3
    & 65.7 & 3.1e-4
    & 71.4 & 5.25e-3
    & 74.4 & 1.50e-3 \\

    \midrule
    \rowcolor{tblRRF}
    RRF (ours)
    & \shortstack[c]{65.7\\{\scriptsize(-2.1\%)}}
    & \shortstack[c]{1.35e-2\\{\scriptsize(-19.2\%)}}
    & \shortstack[c]{78.4\\{\scriptsize(-0.4\%)}}
    & \shortstack[c]{4.5e-4\\{\scriptsize(-71.0\%)}}
    & \shortstack[c]{74.8\\{\scriptsize(0.0\%)}}
    & \shortstack[c]{2.54e-3\\{\scriptsize(-79.7\%)}}
    & \shortstack[c]{80.2\\{\scriptsize(+9.4\%)}}
    & \shortstack[c]{7.1e-4\\{\scriptsize(-72.5\%)}} \\

    \bottomrule
\end{tabular}

\caption{\textbf{Ablation study of the importance ranking strategies.} We compare topology-based signals (degree-only, betweenness-only) and a functional signal (Shapley-only), as well as their fusion via Reciprocal Rank Fusion (RRF), across four representative benchmarks: DROP for reading comprehension and discrete numerical reasoning, MATH and AIME for mathematical and symbolic reasoning (AIME further targeting competition-level problem solving), and MBPP for program synthesis. We report task accuracy (Acc.) and average inference cost per problem (\$/prob.). The percentages in the RRF row indicate the relative change on each benchmark compared to the baseline.}
\label{tab:ablation_importance}
\end{table*}

\begin{figure}[t]
  \centering
  \includegraphics[width=\columnwidth]{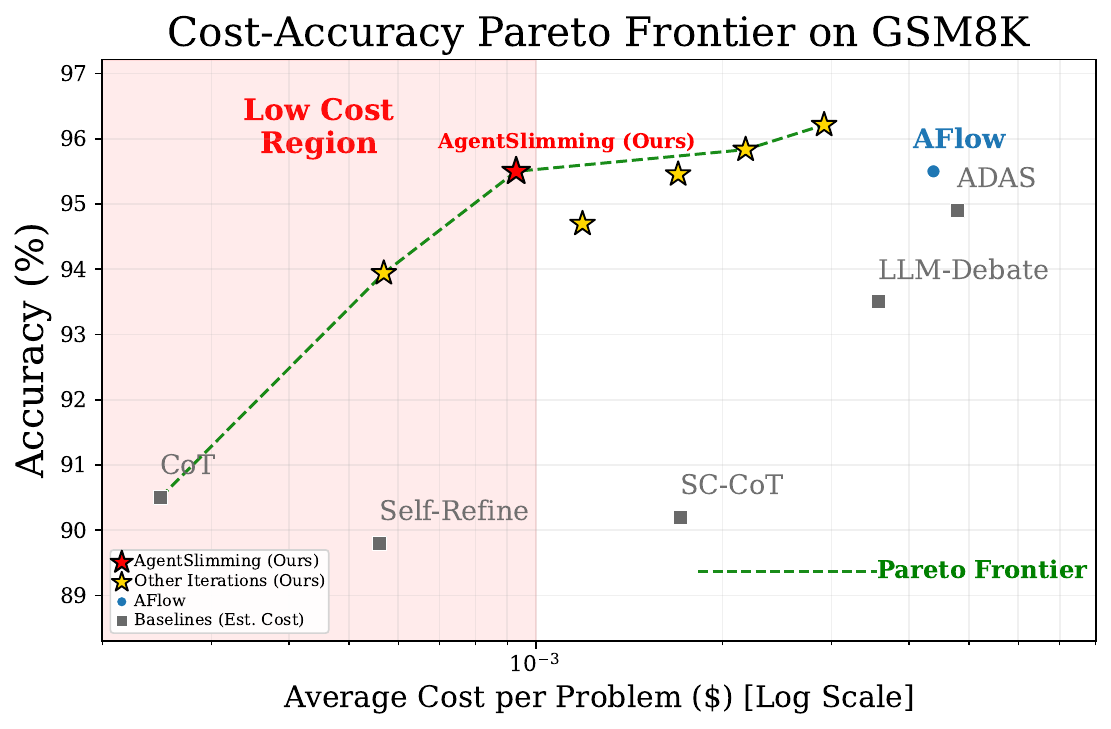}
  \caption{\textbf{Cost-Accuracy Pareto Frontier on the GSM8K dataset.} The x-axis represents the average inference cost per problem (USD), and the y-axis denotes accuracy. Data points for our method correspond to varying iteration rounds. These configurations consistently align with the Pareto frontier, demonstrating optimal cost-quality trade-offs compared to baselines.}
  \label{fig:pareto}
\end{figure}

\subsection{Analysis}
We conducted a comprehensive analysis to identify the sources of efficiency gains in AgentSlimming and validate its practical feasibility.

\begin{figure}[t]
  \centering
  \includegraphics[width=\columnwidth]{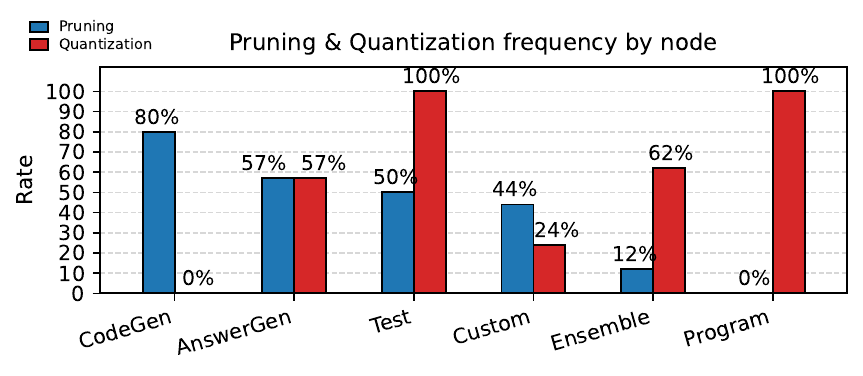}
  \caption{\textbf{Node-type specific compression rates.} We report the ratio of nodes processed by each compression strategy across six categories. Blue bars indicate structural removal (Pruning), while red bars indicate model substitution (Quantization). The variation across categories highlights the adaptive nature of our hybrid importance evaluation.}
  \label{fig:prune_nodetype_frequency}
\end{figure}

\paragraph{Denoising via Pruning} In the initial heavy workflows, we observed redundant parallel generation components, particularly multiple \texttt{CodeGenerate} nodes and \texttt{AnswerGenerate} nodes. Our method consistently deprioritizes these redundant nodes due to their limited marginal benefit, pruning them to yield a simplified graph structure.

\paragraph{Strategic Heterogeneity} When parallel structures remain after pruning, AgentSlimming applies non-uniform compression and retains a mixed-precision configuration. Specifically, it preserves a small number of high-capability \texttt{AnswerGenerate} nodes at full precision and quantizes the remaining parallel auxiliary nodes to lower-cost models.

\paragraph{Cost-Effective Arbitration via Functional Alignment} The quantization frequencies indicate that AgentSlimming primarily compresses nodes whose functionality is procedure-oriented rather than generation-intensive. \texttt{ScEnsembler} and \texttt{Programmer} nodes are infrequently pruned but are frequently quantized, while \texttt{Test} nodes are consistently quantized. Although these nodes can be topologically central, they mainly implement selection, consensus aggregation, and execution-based validation. By selectively quantizing them while retaining high-capability generators, AgentSlimming preserves accuracy and strategically directs the computational budget toward core reasoning rather than auxiliary validation.

\paragraph{Total Evaluation Budget Analysis.} To translate the structural efficiency gains into tangible economic implications, we estimate the total end-to-end evaluation budget for the test sets of each benchmark by multiplying the optimized per-problem cost by the effective sample size. Remarkably, the computational burden is minimal: standard benchmarks like MBPP and GSM8K require less than \textbf{\$2.00} to evaluate the entire test set (\$1.63 and \$1.84, respectively). Even for computationally intensive, high-difficulty benchmarks such as AIME and HotpotQA, the total costs remain strictly affordable at \$26.06 and \$14.76. 

Crucially, this cost-effectiveness also extends beyond inference to the optimization process itself. As a fully training-free framework, AgentSlimming avoids the prohibitive overhead of gradient-based updates and expensive parameter tuning. Furthermore, by leveraging topological priors (e.g., degree and betweenness centrality) for importance scoring, we ensure that pruning and quantization remain computationally efficient and lightweight. To further quantify the trade-off between the one-time optimization overhead and the recurring inference-time savings, we additionally analyze the time-to-break-even of our method, i.e., how many future executions are needed to amortize the search cost through reduced per-query inference cost. We report the detailed formulation and benchmark-wise optimization cost in Appendix~\ref{app:break_even}.

\subsection{Ablation Study}
We conduct two complementary ablation studies. First, we examine the effect of pipeline order by comparing our default \textit{Prune $\rightarrow$ Quantize} design with the reversed \textit{Quantize $\rightarrow$ Prune} variant. As shown in Table~\ref{tab:order_ablation}, reversing the order leads to only marginal changes in the final cost--accuracy trade-off, indicating that the final Pareto frontier is driven mainly by the joint effect of pruning and quantization rather than by a specific execution order. The small differences mainly arise from slightly different search trajectories under the greedy rollback mechanism. We nevertheless adopt \textit{Prune $\rightarrow$ Quantize} in the main framework, because pruning first removes redundant nodes early, shrinks the workflow graph, and reduces the search cost of the subsequent quantization stage.

We then investigate the impact of the node-importance ranking strategy used in pruning and quantization. Our full method employs a weighted Reciprocal Rank Fusion (RRF) scheme to synthesize multiple signals into a unified candidate ranking, whereas the ablated variants rely on a single metric, namely Degree-only, Betweenness-only, and Shapley-only. To isolate the effect of ranking quality from confounders such as search budget, initialization, and stopping criteria, we keep the experimental protocol consistent across stages: for pruning, all variants start from the original workflow; for quantization, each variant inherits the best pruned graph derived from its corresponding pruning strategy and applies the same metric for node selection. Both stages use identical validation subset sizes and baseline-anchored acceptance thresholds. Table~\ref{tab:ablation_importance} summarizes the resulting performance. We observe that single-metric strategies exhibit divergent cost--quality trade-offs across tasks, highlighting their complementary strengths and motivating multi-signal fusion. In contrast, RRF consistently achieves a more robust balance between effectiveness and efficiency. Detailed stage-wise ablation results are provided in Appendix~\ref{app:stage-wise}.

\subsection{Case Study}
\begin{figure}[t]
  \centering
  \includegraphics[width=\columnwidth]{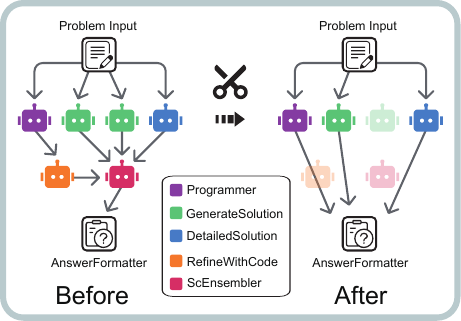}
  \caption{\textbf{Visualizing the pruning process of MATH workflow.} (\textit{Before}) The initial workflow features complex parallel reasoning paths and an ensemble mechanism. (\textit{After}) The streamlined workflow after AgentSlimming's pruning stage. It effectively identifies the core reasoning backbone by pruning redundant parallel paths and refinement steps, thereby reducing cost.}
  \label{fig:case_study}
  \vspace{-0.3cm}
\end{figure}

We analyze the pruning process of the workflow on MATH to illustrate structural simplification. Starting from a complex graph with parallel reasoning paths and refinement steps, AgentSlimming identified redundancy in the dual \texttt{AnswerGenerate} nodes and the marginal utility of the \texttt{CodeRefine} node. Concretely, probe-based importance ranking suggested that one generator path consistently dominates in contribution, while refinement rarely corrects errors relative to its token cost. Notably, pruning the parallel path diminished the utility of the downstream \texttt{ScEnsemble} node, leading to its removal (as demonstrated in Figure~\ref{fig:case_study}). The result is a streamlined pipeline that preserves core reasoning logic while achieving substantial cost reductions, empirically validating our topo-functional optimization strategy. The complete details of this case study are provided in Appendix~\ref{app:case}.

\section{Conclusion}
We introduced AgentSlimming, a unified framework that optimizes multi-agent workflows via structural pruning and semantic quantization. Guided by a novel hybrid metric combining topological and functional signals, AgentSlimming achieves up to \textbf{78.9\%} cost reduction across diverse benchmarks. Importantly, this efficiency is attained with negligible impact on reasoning quality, matching or even exceeding baseline performance. Our method shows that sparse, heterogeneous topologies can effectively replace computationally redundant dense multi-agent systems, establishing a new Pareto frontier for scalable agentic collaboration.

\section*{Limitations}
    We evaluate GPT-4.1-series and Qwen3-series models, but do not benchmark newer frontier models. Like AFlow, our framework currently operates at the task level to ensure fair baseline comparisons. For future work, we plan to explore instance-level dynamic adaptation, utilizing multi-agent memory to adjust workflows and compute based on query difficulty.

\section*{Acknowledgements}
This work was supported by Alibaba Group through Alibaba Innovative Research Program.

\section*{Ethics Statement}
We acknowledge that all authors are informed about and adhere to the ACL ARR Code of Ethics and the Code of Conduct.

\paragraph{Risks}
Our benchmarks are sourced from publicly available datasets. We cannot guarantee that they are free of socially harmful, biased, or toxic content. In addition, AgentSlimming optimizes cost by pruning and replacing nodes, which may alter the behavior of the original workflow in unexpected ways; when applied to safety-critical domains, such changes could amplify errors or reduce reliability. 
We used LLM-based tools only for grammar and language polishing; all technical content, experiments, and claims were written and verified by the authors.

\bibliography{custom}

\clearpage
\appendix

\section*{Appendix}

\setcounter{paragraph}{0}
\renewcommand{\theparagraph}{}

% \clearpage
\section{Algorithm}
\begin{algorithm}[!htbp]
% \small
\caption{Prune--Quantize--Fine-tune Pipeline}
\label{alg:mpqr}
\begin{algorithmic}[1]
\STATE \textbf{Input:} initial graph $\mathcal{G}_{\mathrm{base}}$, validation set $\mathcal{D}_{\mathrm{val}}$, probe size $m$, budget $B$.
\STATE \hspace{1.2em} Top-$k$ size $k$, thresholds $\tau_p,\tau_q$.
\STATE \hspace{1.2em} RRF weights $\mathbf{w}$, low-cost mapping $\pi$.
\STATE \textbf{Output:} optimized graph $\mathcal{G}^\star$.
\STATE Initialize $\mathcal{G} \leftarrow \mathcal{G}_{\mathrm{base}}$.
\STATE \textbf{Step 1: AgentPruner.}
\STATE Set baseline $S_0 \leftarrow S(\mathcal{G})$.
\WHILE{pruning budget remains}
  \STATE Rank nodes by probe signals; fuse by weighted RRF; get Top-$k$ $\{v_1,v_2,\ldots,v_k\}$.
  \FOR{$v\in\{v_1,v_2,\ldots,v_k\}$}
    \STATE $\mathcal{G}_p \leftarrow \mathrm{Surgery}(\mathcal{G},v)$; compute $S(\mathcal{G}_p)$.
    \IF{$S(\mathcal{G}_p) \ge \tau_p \cdot S_0$} \STATE $\mathcal{G} \leftarrow \mathcal{G}_p$; \textbf{break} \ENDIF
  \ENDFOR
  \STATE Stop if no candidate is accepted.
\ENDWHILE

\STATE \textbf{Step 2: AgentQuant.}
\STATE Set baseline $S_0 \leftarrow S(\mathcal{G})$.
\WHILE{quantization budget remains}
  \STATE Rank LLM nodes by probe replacement signals; fuse by weighted RRF; get Top-$k$ $\{v_1,v_2,\ldots,v_k\}$.
  \FOR{$v\in\{v_1,v_2,\ldots,v_k\}$}
    \STATE $\mathcal{G}_q \leftarrow \mathrm{Replace}(\mathcal{G},v,\pi(v))$; compute $S(\mathcal{G}_q)$.
    \IF{$S(\mathcal{G}_q) \ge \tau_q \cdot S_0$} \STATE $\mathcal{G} \leftarrow \mathcal{G}_q$; \textbf{break} \ENDIF
  \ENDFOR
  \STATE Stop if no candidate is accepted.
\ENDWHILE

\STATE \textbf{Step 3: AgentTuner.}
  \STATE Run MCTS on $\mathcal{G}$ with atomic edits; update $\mathcal{G}$.

\STATE \textbf{return} $\mathcal{G}^\star \leftarrow \arg\max_{\mathcal{H}\in\mathcal{K}} (S(\mathcal{H}),-C(\mathcal{H}))$.
\item[] \phantom{\textbf{return}} \textbf{s.t.} $C(\mathcal{G}^\star)\le B$.
\end{algorithmic}
\end{algorithm}

\section{Case Study}
\label{app:case}
\subsection{Workflow Overview and Legend Mapping}
Figure~\ref{fig:case_study} visualizes the pruning process on the redundant workflow graph generated on MATH \citep{dan2021math}.
To match the legend used in the figure, we map the implementation node IDs to the paper terms as follows: (1) \texttt{GenerateSolutionA} / \texttt{GenerateSolutionB} correspond to dual \texttt{AnswerGenerate} nodes. (2) \texttt{RefineWithCode} corresponds to the \texttt{CodeRefine} node. (3) \texttt{ScEnsembler} corresponds to the \texttt{ScEnsemble} node for aggregation. (4) \texttt{AnswerFormatter} formats the final output to satisfy benchmark constraints.

\subsection{Pruning Rationale and Structural Simplification}
The original workflow (Figure~\ref{fig:case_study}, left) contains multiple parallel reasoning paths: a
program-execution path (\texttt{Programmer} $\rightarrow$ \texttt{RefineWithCode}), a detailed chain-of-thought path
(\texttt{DetailedSolution}), and two additional solution generation paths
(\texttt{GenerateSolutionA} and \texttt{GenerateSolutionB}). All candidate answers are routed into
\texttt{ScEnsembler} for self-consistency selection before being formatted by \texttt{AnswerFormatter}. AgentSlimming identifies two main sources of redundancy: (1) The two \texttt{AnswerGenerate} branches are highly overlapping; keeping one branch provides most of the diversity gain. (2) The marginal utility of the \texttt{CodeRefine} step is limited for this workflow; its benefits do not justify the additional token cost.

\section{Experiment}
\label{app:details}
\subsection{Baselines}
To ensure a fair and rigorous comparison that matches the scale of our optimized workflows, we configured the LLM-Debate baseline with \textbf{4 agents and 2 debate rounds}. This specific hyperparameter setting aligns the computational complexity and graph scale of the baseline with our method.

\subsection{Hyperparameters}
We report the key hyperparameters used across all experiments. The acceptance thresholds are set to $\tau_p$ = $\tau_q$ = $0.95$, where higher values enforce more conservative selection and typically yield faster convergence, while lower values permit broader exploration. The top-$k$ candidate pool size is set to k = 3 for both pruning and quantization evaluation. The RRF smoothing constant is set to $\kappa = \max(10, |V|)$. For the RRF importance weights, we set $w_{\text{degree}} = 1$, $w_{\text{betweenness}} = 1$, $w_{\text{shapley}} = 2$, and $w_{\Delta\text{cost}} = 1$, where the Shapley weight is assigned a higher value to emphasize cooperative contribution signals.

\subsection{Nodes}
Following AFlow, we treat an operator as a node in the DAG: each node corresponds to an executable operator (typically an LLM call) with fixed I/O constraints, including its own prompt, available resources/tools, and input–output schema. We provide complete per-node definitions and the default prompts for standard nodes in Section~\ref{app:nodes}.

\subsection{Source Code}
We provide the executable code for the initial workflow graphs of MATH, MBPP, and HotpotQA in Section~\ref{app:graphs} to ensure reproducibility.

\section{Stage-wise Contribution Details}
\label{app:stage-wise}

In the AgentTuner stage, we adopt an offline workflow search procedure inspired by Monte Carlo Tree Search (MCTS). Each node in the search space corresponds to a complete workflow \(W\), i.e., a full multi-agent system. At each iteration, the algorithm first selects a parent workflow from the historical candidate pool, and then invokes the LLM optimizer to generate a local modification based on both prior search experience and the operator set \(O\). Such modifications may include prompt refinement, rewiring node connections, or recomposing operators, each yielding a new workflow candidate.

\begin{table*}[!t]
\centering
\scriptsize
\setlength{\tabcolsep}{9pt}
\renewcommand{\arraystretch}{1.08}

\begin{tabular}{lcccccccc}
    \toprule
    \rowcolor{tblhead}
    \textbf{Method}
    & \multicolumn{2}{c}{\textbf{MATH}}
    & \multicolumn{2}{c}{\textbf{AIME}}
    & \multicolumn{2}{c}{\textbf{GSM8K}}
    & \multicolumn{2}{c}{\textbf{HotpotQA}} \\
    \cmidrule(lr){2-3} \cmidrule(lr){4-5} \cmidrule(lr){6-7} \cmidrule(lr){8-9}
    & Score & Cost (\$/prob.)
    & Score & Cost (\$/prob.)
    & Score & Cost (\$/prob.)
    & Score & Cost (\$/prob.) \\
    \midrule

    Original
    & 73.1 & 1.03e-2
    & 66.4 & 1.55e-2
    & 96.6 & 4.34e-3
    & 77.0 & 7.65e-3 \\

    After pruning
    & 75.6 & 3.30e-3
    & 67.1 & 1.19e-2
    & 95.8 & 2.18e-3
    & 76.7 & 5.55e-3 \\

    After quantization
    & 74.8 & 2.54e-3
    & 67.1 & 1.19e-2
    & 93.9 & 5.70e-4
    & 72.8 & 2.90e-3 \\

    After tuning
    & 75.6 & 1.04e-2
    & 69.6 & 1.28e-1
    & 95.1 & 6.10e-4
    & 71.9 & 4.20e-3 \\

    \midrule
    \rowcolor{tblsubhead}
    \textbf{Method}
    & \multicolumn{2}{c}{\textbf{MBPP}}
    & \multicolumn{2}{c}{\textbf{DROP}}
    & \multicolumn{2}{c}{\textbf{LiveCode}}
    & \multicolumn{2}{c}{\textbf{MusiqueAns}} \\
    \cmidrule(lr){2-3} \cmidrule(lr){4-5} \cmidrule(lr){6-7} \cmidrule(lr){8-9}
    & Score & Cost (\$/prob.)
    & Score & Cost (\$/prob.)
    & Score & Cost (\$/prob.)
    & Score & Cost (\$/prob.) \\
    \midrule

    Original
    & 69.8 & 2.57e-3
    & 77.9 & 1.55e-3
    & 60.2 & 1.09e-2
    & 85.3 & 1.07e-2 \\

    After pruning
    & 79.7 & 1.63e-3
    & 77.7 & 6.40e-4
    & 59.8 & 4.55e-3
    & 86.4 & 3.51e-3 \\

    After quantization
    & 79.2 & 7.10e-4
    & 78.4 & 4.50e-4
    & 61.4 & 3.20e-3
    & 85.3 & 9.90e-4 \\

    After tuning
    & 80.2 & 7.10e-4
    & 83.3 & 9.50e-4
    & 61.7 & 2.59e-3
    & 85.8 & 9.90e-4 \\

    \bottomrule
\end{tabular}

\caption{\textbf{Stage-wise contribution analysis of pruning, quantization, and AgentTuner.} We report the score and average inference cost per problem at different stages of the optimization pipeline across eight representative benchmarks. The results show that pruning alone already yields substantial cost reduction and can even improve task performance on several datasets. Quantization further reduces cost, while sometimes introducing a moderate accuracy drop, reflecting the standard compression trade-off. AgentTuner serves as an optional offline enhancement step that may further improve performance on some tasks, but it is not required for obtaining strong cost efficiency, nor does it always provide the best overall cost--accuracy trade-off.}
\label{tab:stagewise_contrib}
\end{table*}

Different from classical UCT-style tree policies, our parent selection does not follow an explicit upper-confidence bound. Instead, we employ a soft mixed probability mechanism that interpolates between uniform exploration and score-based exploitation. Concretely, for \(n\) candidate parent workflows, the probability of selecting the \(i\)-th candidate is defined as:
\begingroup
\small
\begin{equation}
\begin{aligned}
P_{\text{mixed}}(i)
&=\lambda\cdot \frac{1}{n} \\
&\quad +(1-\lambda)\cdot
\frac{\exp\!\big(\alpha (s_i-s_{\max})\big)}
{\sum_{j=1}^{n}\exp\!\big(\alpha (s_j-s_{\max})\big)}
\end{aligned}
\end{equation}
\endgroup
where \(s_i\) denotes the average validation score of workflow \(i\) on the validation set \(\mathcal{D}_\mathrm{val}\), \(s_{\max}\) is the maximum score among the current candidates, \(\lambda\) controls the exploration strength, and \(\alpha\) controls the preference toward high-performing candidates. This design encourages broad exploration in the early stage while gradually biasing the search toward stronger workflows.

For each newly generated workflow, we evaluate its \((\text{score}, \text{cost})\) on \(\mathcal{D}_\mathrm{val}\). To reduce evaluation noise, the workflow can be executed multiple times and the average performance is recorded. The reward fed back to the search process is defined as the average validation score:
\begin{equation}
    r(W)=\mathrm{avgScore}(W;D_V)
\end{equation}
The search is equipped with an early stopping rule: the procedure terminates when the top-$k$ candidates remain unchanged for $n$ consecutive rounds, or equivalently when the mean score of the top-$k$ set no longer improves. The full hyperparameter configurations are as follows: $\lambda$ = 0.3, $\alpha$ = 0.2, $k$ = 3, stopping patience = 5, and the number of repeated evaluations = 1.

It is important to clarify the role of AgentTuner in our framework. The core contribution of this work lies in the structural optimization pipeline---namely, pruning and quantization---together with the RRF-based fusion of node-importance signals. AgentTuner is an optional offline enhancement module that can be applied when additional search budget is available, with the goal of further exploring the accuracy--cost Pareto frontier. It is not required for the validity of the proposed method, nor is it specifically designed to compensate for possible performance loss introduced by pruning or quantization.

The stage-wise results in Table~\ref{tab:stagewise_contrib} further support this interpretation. The pruning stage alone already delivers substantial cost reduction, and even improving accuracy on several datasets, indicating that structural simplification does not inherently rely on sacrificing task performance. The quantization stage yields additional cost savings, although it may incur moderate accuracy degradation on some tasks, which is consistent with the standard compression trade-off. Importantly, even without AgentTuner, the workflow obtained after pruning and quantization already achieves strong cost efficiency with acceptable performance, demonstrating that AgentTuner is not a mandatory recovery step for restoring model quality.

Moreover, AgentTuner does not always provide the best overall utility. On some datasets, it improves accuracy but also introduces a noticeable increase in inference cost. This observation further confirms that AgentTuner should be viewed as an optional module rather than a necessary corrective component of the system. In practice, it is most beneficial in scenarios where high accuracy is prioritized over cost, where compression causes a non-trivial accuracy drop and partial recovery is desired, or where sufficient budget is available to search for solutions closer to the Pareto-optimal frontier.

\section{Break-even Analysis of Optimization Cost}
\label{app:break_even}

To quantify the trade-off between the one-time optimization overhead and the recurring inference-time savings, we provide a time-to-break-even analysis. Let \(C_{\text{optimize}}\) denote the total cost incurred during the structural optimization/search stage. Let \(C_{\text{base}}\) and \(C_{\text{ours}}\) denote the average API cost per inference before optimization (the baseline method) and after optimization (our method), respectively. Then, the per-inference cost saving is defined as:
\begin{equation}
    \Delta C = C_{\text{base}} - C_{\text{ours}}
\end{equation}
Accordingly, the break-even point is defined as the number of future executions required to amortize the one-time optimization cost:
\begin{equation}
    N_{\text{break-even}} = \frac{C_{\text{optimize}}}{\Delta C}
\end{equation}
\begin{table}[t]
\centering
\scriptsize
\setlength{\tabcolsep}{6pt}
\renewcommand{\arraystretch}{1.08}
\begin{tabular}{@{}lcc@{}}
    \toprule
    \textbf{Datasets} & \textbf{Total optimization cost (\$)} & \textbf{Break-even inference calls} \\
    \midrule
    AIME       & 43.07 & 13459 \\
    MATH       & 14.04 & 2742  \\
    DROP       & 2.26  & 3767  \\
    MBPP       & 2.26  & 1222  \\
    GSM8K      & 5.70  & 1652  \\
    HotpotQA   & 18.57 & 8889  \\
    LiveCode   & 26.52 & 2873  \\
    MusiqueAns & 25.85 & 10508 \\
    \bottomrule
\end{tabular}
\caption{\textbf{Break-even analysis of optimization cost.} We report the total one-time optimization cost \(C_{\text{optimize}}\) and the corresponding number of future executions required to amortize this cost relative to AFlow.}
\label{tab:break_even}
\end{table}

Here, \(C_{\text{optimize}}\) is paid only once during the optimization stage, whereas \(\Delta C\) yields recurring savings in every subsequent deployment. Therefore, as the optimized workflow is executed repeatedly, the upfront search cost is gradually amortized over time.

Table~\ref{tab:break_even} reports the benchmark-wise \(C_{\text{optimize}}\) and the resulting \(N_{\text{break-even}}\) values relative to AFlow, providing a more complete view of the trade-off between search overhead and inference-time savings.

\begin{table}[!t]
\centering
\small
\setlength{\tabcolsep}{9pt}
\renewcommand{\arraystretch}{1.12}
\begin{tabular}{lcc}
    \toprule
    \textbf{Probe Size} & \multicolumn{2}{c}{\textbf{HotpotQA}} \\
    \cmidrule(lr){2-3}
    & \textbf{Acc.} & \textbf{Cost (\$/prob.)} \\
    \midrule
    10  & 75.6 & 5.89e-3 \\
    50  & 77.0 & 5.55e-3 \\
    100 & 75.3 & 3.79e-3 \\
    \bottomrule
\end{tabular}
\caption{\textbf{Sensitivity analysis of probe dataset size on HotpotQA.} We vary the probe size used for node-importance evaluation and report the resulting task accuracy and average inference cost per problem.}
\label{tab:probe_size}
\end{table}

\begin{table}[!t]
\centering
\small
\setlength{\tabcolsep}{4pt}
\renewcommand{\arraystretch}{1.10}
\begin{tabular}{lcccc}
    \toprule
    \textbf{Method}
    & \multicolumn{2}{c}{\textbf{MATH}}
    & \multicolumn{2}{c}{\textbf{HotpotQA}} \\
    \cmidrule(lr){2-3} \cmidrule(lr){4-5}
    & \textbf{Acc.} & \textbf{Cost}
    & \textbf{Acc.} & \textbf{Cost} \\
    \midrule
    Baseline (ADAS)
    & 69.0 & 1.15e-2
    & 81.6 & 5.89e-3 \\
    Ours
    & 70.4 & 9.67e-3
    & 79.4 & 2.75e-3 \\
    \bottomrule
\end{tabular}
\caption{\textbf{Experiments on ADAS-based workflows.} We apply AgentSlimming to workflows derived from ADAS without changing the core optimization algorithm or hyperparameter settings.}
\label{tab:adas_generalization}
\end{table}

\begin{table*}[!t]
\centering
\small
\setlength{\tabcolsep}{4pt}
\renewcommand{\arraystretch}{1.12}
\begin{tabular}{lcccccccc}
    \toprule
    \textbf{Method} 
    & \multicolumn{2}{c}{\textbf{AIME}} 
    & \multicolumn{2}{c}{\textbf{HotpotQA}} 
    & \multicolumn{2}{c}{\textbf{MBPP}} 
    & \multicolumn{2}{c}{\textbf{GSM8K}} \\
    \cmidrule(lr){2-3} \cmidrule(lr){4-5} \cmidrule(lr){6-7} \cmidrule(lr){8-9}
    & \textbf{Acc.} & \textbf{Cost (\$/prob.)}
    & \textbf{Acc.} & \textbf{Cost (\$/prob.)}
    & \textbf{Acc.} & \textbf{Cost (\$/prob.)}
    & \textbf{Acc.} & \textbf{Cost (\$/prob.)} \\
    \midrule
    Baseline (AFlow) 
    & 87.1 & 1.19e-2 
    & 81.6 & 5.89e-3 
    & 73.3 & 2.84e-3 
    & 96.6 & 3.04e-3 \\
    Ours 
    & 87.9 & 6.59e-3 
    & 79.4 & 2.76e-3 
    & 75.6 & 1.06e-3 
    & 97.9 & 5.57e-4 \\
    \bottomrule
\end{tabular}
\caption{\textbf{Cross-architecture validation on the Qwen3 series.} We replace the high-/low-cost nodes with \texttt{qwen3-235b-a22b-instruct-2507} and \texttt{qwen3-30b-a3b-instruct-2507}, respectively, while keeping all other algorithms and hyperparameters unchanged. Across AIME, HotpotQA, MBPP, and GSM8K, AgentSlimming consistently achieves substantial cost reductions while maintaining competitive overall performance.}
\label{tab:qwen3}
\end{table*}
\begin{table*}[!t]
\centering
\small
\setlength{\tabcolsep}{8pt}
\renewcommand{\arraystretch}{1.12}
\begin{tabular}{lcccc}
    \toprule
    \textbf{Method}
    & \multicolumn{2}{c}{\textbf{MATH}}
    & \multicolumn{2}{c}{\textbf{GSM8K}} \\
    \cmidrule(lr){2-3} \cmidrule(lr){4-5}
    & \textbf{Acc.} & \textbf{Cost (\$/prob.)}
    & \textbf{Acc.} & \textbf{Cost (\$/prob.)} \\
    \midrule
    AFlow
    & 74.8 & 1.20e-2
    & 95.5 & 4.38e-3 \\
    
    Ours 
    & 73.9 & 6.88e-3
    & 95.5 & 9.30e-4 \\

    Cross-dataset results 
    & 77.3 & 6.63e-3
    & 96.6 & 2.63e-3 \\
    \bottomrule
\end{tabular}
\caption{\textbf{Cross-dataset validation between GSM8K and MATH.} We optimize the workflow structure on one dataset and directly evaluate the resulting structure on the other, without changing the optimization algorithm or hyperparameter settings. The transferred structures remain competitive in both accuracy and cost efficiency, suggesting that AgentSlimming exhibits non-trivial cross-dataset generalization across same-type reasoning tasks.}
\label{tab:cross_dataset}
\end{table*}

\section{Probe Dataset Size Sensitivity}
\label{app:probe_size}
To evaluate the sensitivity of our method to the probe dataset size, we conduct additional experiments on HotpotQA with probe sizes of 10, 50, and 100. The results are summarized in Table~\ref{tab:probe_size}. We observe that even a moderate probe size of 50 is sufficient to produce a stable node-importance ranking and achieves the best accuracy among the three settings. Further increasing the probe size to 100 reduces inference cost, but also leads to a drop in accuracy. Empirically, a probe size in the range of 30--60 already provides strong performance, and the importance scores stabilize with as few as 50 probe samples. This suggests that our optimization process does not require a large probe dataset to obtain consistent pruning decisions, which further improves the practicality and efficiency of the overall framework.

\section{Additional Generalization Studies}
\label{app:generalization}

\subsection{Experiments on the Qwen3 Series}
\label{app:qwen3}

To evaluate whether our method generalizes across model families, we further conduct cross-architecture validation on the Qwen3 series. Specifically, we replace the high-/low-cost nodes with \texttt{qwen3-235b-a22b-instruct-2507} and \texttt{qwen3-30b-a3b-instruct-2507}, respectively, while keeping all other algorithms and hyperparameters unchanged. We evaluate on four datasets: AIME, HotpotQA, MBPP, and GSM8K, which cover both high- and low-difficulty settings as well as diverse task types.

The results are summarized in Table~\ref{tab:qwen3}. We observe that AgentSlimming still delivers substantial cost reductions under this cross-architecture setting, while preserving competitive overall accuracy. In particular, our method improves accuracy on AIME, MBPP, and GSM8K, while incurring only a moderate drop on HotpotQA. These results suggest that the effectiveness of AgentSlimming is not tied to a specific model family, and that its cost-saving benefits transfer well to the Qwen3 series.

\subsection{Cross-Dataset Validation between GSM8K and MATH}
\label{app:cross_dataset}

To further address concerns about generalization to unseen data, we conduct additional cross-dataset validation between GSM8K and MATH. Specifically, we optimize the workflow structure on one dataset and then directly evaluate the resulting structure on the other dataset. No additional structural search is performed on the target dataset, and all optimization algorithms and hyperparameter settings are kept unchanged.

Table~\ref{tab:cross_dataset} presents the cross-dataset results. These results suggest that the optimized structures discovered by AgentSlimming are not overly specialized to a single dataset. Instead, they transfer reasonably well across same-type reasoning datasets, indicating non-trivial cross-dataset generalization ability.

\subsection{Experiments on ADAS-Based Workflows}
\label{app:adas}

To further verify that AgentSlimming does not rely on AFlow-specific implementations, we conduct an additional cross-framework transfer study by applying the same slimming pipeline to ADAS, an MAS framework that differs from AFlow in both its optimization mechanism and overall system organization, under the same tasks and evaluation protocol. We keep the core optimization algorithm and hyperparameter settings unchanged, and evaluate the resulting compressed workflows on MATH and HotpotQA.

The results are summarized in Table~\ref{tab:adas_generalization}. We observe that AgentSlimming remains effective on ADAS-based workflows, achieving substantial cost reduction while keeping performance degradation within an acceptable range, consistent with the trend observed on AFlow. These results support the robustness and transferability of AgentSlimming beyond a single workflow family.

More broadly, AgentSlimming only assumes that the underlying multi-agent workflow can be modeled as a DAG. Under this abstraction, we estimate an importance score for each agent node and iteratively perform removal or low-cost replacement, with a baseline-anchored acceptance rule to prevent performance collapse. Therefore, AgentSlimming can be used as a plug-and-play module for a broad range of graph-structured MAS, as long as the workflow can be abstracted into nodes and dependencies and the system provides a stable task-level evaluation signal. We also note that systems that cannot be reasonably expressed as a DAG, or that lack a stable evaluation signal, may require additional graph transformation or evaluation design.

\begin{onecolumn}
\section{Nodes}
\label{app:nodes}
\subsection{Definitions of Standard Nodes}
    \begin{tcolorbox}[rounded corners, breakable, colframe=black!70, colback=white, boxrule=4pt, boxsep=0.5pt, enhanced, shadow={3pt}{-3pt}{0pt}{opacity=0.3}]
    
    \centering
    \begin{tabularx}{\textwidth}{>{\raggedright\arraybackslash}p{2.7cm} >{\raggedright\arraybackslash}p{2.8cm} >{\raggedright\arraybackslash}X >{\raggedright\arraybackslash}X}
    \textbf{Node name} & \textbf{Role} & \textbf{Functionality} & \textbf{Applicable datasets} \\
    \hline
    AnswerGenerate Node & Answer Generator & Directly generates an answer to the input question using an LLM. It is the standard node for direct question answering. & General QA, Reasoning (e.g., GSM8K, MATH, HotpotQA) \\
    \hline
    Programmer Node & Program Executor & Generates Python code (PoT/PAL style) to solve the problem and executes it in a local sandbox. It includes a retry mechanism for execution failures. & Math \& Logic tasks requiring calculation (e.g., MATH, GSM8K) \\
    \hline
    CustomCode-Generate Node & Code Generator & Focuses solely on generating executable Python code (e.g., a solve() function) based on the problem, without executing the code immediately. & Code Generation (e.g., LiveCode, MBPP) \\
    \hline
    Test Node & Self-Corrector & Executes generated code and iteratively refines/fixes it based on execution feedback (runtime errors or incorrect outputs). & Complex Reasoning \& Coding (e.g., LiveCode, MBPP) \\
    \hline
    ScEnsemble Node & Ensemble Selector & Aggregates multiple candidate solutions from upstream nodes and selects the best one using Self-Consistency or majority voting logic. & General Reasoning (e.g., GSM8K, MATH) \\
    \hline
    CodeRefine Node & Code Refiner & Refines and improves an existing draft solution by leveraging programmatic reasoning or code-based feedback before final aggregation/formatting. & Math \& Multi-step Reasoning with program-aided refinement (e.g., GSM8K, MATH, AIME) \\
    \hline
    Custom Node & Custom Operator & Executes a user-defined prompt or custom logic, providing flexibility for specific sub-tasks not covered by standard nodes. & Any / Agnostic \\
    \end{tabularx}
    \end{tcolorbox}
\end{onecolumn}

\begin{onecolumn}
\subsection{Default Prompts for Standard Nodes}
    \begin{tcolorbox}[rounded corners, breakable, colframe=black!70, colback=white, boxrule=4pt, boxsep=0.5pt, enhanced, shadow={3pt}{-3pt}{0pt}{opacity=0.3}, title={Prompt of Standard Nodes}]
    \label{lis:prompt}
        \begin{lstlisting}[breaklines=true,showstringspaces=false,language=python]
        
        PROMPT_ANSWERGENERATOR = """
            Think step by step and solve the problem.
            1. In the "thought" field, explain your thinking process in detail.
            2. In the "answer" field, provide the final answer concisely and clearly. The answer should be a direct response to the question, without including explanations or reasoning.
            
            Your task: {input}
        """
        
        
        PROMPT_CUSTOMCODEGENERATE = """
            You are a professional Python programmer. Your task is to write complete, self-contained code based on a given problem and output the answer. The code should include all necessary imports and dependencies, and be ready to run without additional setup or environment configuration.
        
            Problem: {problem}
            Resources: {input}
        
            Your code should:
            1. Implement the calculation steps described in the problem.
            2. Define a function named `{function_name}` that performs the calculation and returns the result.
            3. Return only runnable Python code without explanations.
        """
        
        
        PROMPT_SCENSEMBLE = """
            Given the problem described as follows: {problem}
            Several candidate solutions have been generated to address the given problem. They are as follows:
            {solutions}
        
            Your task is to act as an expert evaluator. Carefully evaluate these solutions and identify the definitive answer that appears most frequently across them (Majority Consensus).
            Analyze the solutions to determine the most consistent and reliable outcome.
        """
        
        
        PROMPT_PROGRAMMER = """
            You are a professional Python programmer. Your task is to write complete, self-contained code based on a given problem and output the answer. The code should include all necessary imports and dependencies, and be ready to run without additional setup or environment configuration.
            
            Problem: {problem}
            Resources: {input}
        
            Your code should:
            1. Implement the calculation steps described in the problem.
            2. Define a function named `solve` that performs the calculation and returns the result. The `solve` function should not require any input parameters; instead, it should obtain all necessary inputs from within the function or from globally defined variables.
            3. `solve` function return the final calculation result.
        
            Please ensure your code is efficient, well-commented, and follows Python best practices. The output should be limited to basic data types such as strings, integers, and floats. It is prohibited to transmit images or other file formats. The code output is intended for a text-based language model.
        """
        
        PROMPT_TEST = """
            Given a problem and a python code solution which failed to pass test or execute, you need to analyze the reason for the failure and propose a better code solution.
            
            Problem: {problem}
        
            Failure details:
            {error_info}
        
            Please provide a "reflection" field explaining the failed test cases and code solution, and a "solution" field containing a better code solution without any additional text or test cases.
        """
        
        
        PROMPT_ANSWERFORMAT = """
            You are an answer formatter for the {dataset_name} dataset.
            
            Task Context:
            {task_context}
        
            Format Requirements:
            {format_requirements}
        
            Original Answer:
            {original_answer}
        
            Return only the final formatted answer according to the requirements.
        """
        \end{lstlisting}
    \end{tcolorbox}

\section{Examples of Initial Workflow Graphs}
\label{app:graphs}
\subsection{Minimal initial workflow graph on MATH dataset}
    \begin{tcolorbox}[rounded corners, breakable, colframe=black!70, colback=white, boxrule=4pt, boxsep=0.5pt, enhanced, shadow={3pt}{-3pt}{0pt}{opacity=0.3}]
        \label{lis:graph}
        \begin{lstlisting}[breaklines=true,showstringspaces=false,language=python]
        from src.core.edge import Edge
        from src.core.graphflow import GraphFlow
        from src.core.nodes.answer_format_node import AnswerFormatNode
        from src.core.nodes.custom_node import CustomNode
        from src.core.nodes.input_node import InputNode
        from src.core.workflow import BaseWorkflow
        from . import prompt as prompt_custom
        
        class Workflow(BaseWorkflow):
            def _build_graph(self) -> GraphFlow:
                Input = InputNode(
                    node_id="Input",
                    node_llm_config=self.llm_config,
                    description="Graph input entry.",
                )
                Reasoner = CustomNode(
                    node_id="Reasoner",
                    node_prompt=prompt_custom.PROMPT_REASONING,
                    node_llm_config=self.llm_config,
                    description="Reason over the problem and produce a concise answer.",
                )
                AnswerFormatter = AnswerFormatNode(
                    node_id="AnswerFormatter",
                    dataset_name=self.dataset,
                    node_llm_config=self.llm_config,
                    description="Format the final answer for the target dataset.",
                )
                return GraphFlow(
                    nodes=[Input, Reasoner, AnswerFormatter],
                    edges=[
                        Edge(source="Input", target="Reasoner"),
                        Edge(source="Reasoner", target="AnswerFormatter"),
                    ],
                    entry_node_ids=["Input"],
                    final_node_id="AnswerFormatter",
                    description="Minimal MATH round_1 workflow.",
                )
        \end{lstlisting}
    \end{tcolorbox}
    
    \begin{tcolorbox}[rounded corners, breakable, colframe=black!70, colback=white, boxrule=4pt, boxsep=0.5pt, enhanced, shadow={3pt}{-3pt}{0pt}{opacity=0.3}, title={Prompt of a \textit{Reasoner} Custom Node}]
        \begin{lstlisting}[breaklines=true,showstringspaces=false,language=python]
        PROMPT_REASONING = (
            "Solve the problem carefully. Identify the key facts, reason step by step, and return a concise answer."
        )
        \end{lstlisting}
    \end{tcolorbox}

\subsection{Minimal initial workflow graph on MBPP dataset}
    \begin{tcolorbox}[rounded corners, breakable, colframe=black!70, colback=white, boxrule=4pt, boxsep=0.5pt, enhanced, shadow={3pt}{-3pt}{0pt}{opacity=0.3}]
        \begin{lstlisting}[breaklines=true,showstringspaces=false,language=python]
        from src.core.edge import Edge
        from src.core.graphflow import GraphFlow
        from src.core.nodes.answer_format_node import AnswerFormatNode
        from src.core.nodes.custom_code_generate_node import CustomCodeGenerateNode
        from src.core.nodes.input_node import InputNode
        from src.core.workflow import BaseWorkflow
        
        class Workflow(BaseWorkflow):
            def _build_graph(self) -> GraphFlow:
                Input = InputNode(
                    node_id="Input",
                    node_llm_config=self.llm_config,
                    description="Graph input entry.",
                )
                CustomCodeGenerator = CustomCodeGenerateNode(
                    node_id="CustomCodeGenerator",
                    node_llm_config=self.llm_config,
                    description="Generates Python code for code-generation benchmarks and preserves the entry point.",
                )
                AnswerFormatter = AnswerFormatNode(
                    node_id="AnswerFormatter",
                    dataset_name=self.dataset,
                    node_llm_config=self.llm_config,
                    description="Format the final answer for the target dataset.",
                )
                return GraphFlow(
                    nodes=[Input, CustomCodeGenerator, AnswerFormatter],
                    edges=[
                        Edge(source="Input", target="CustomCodeGenerator"),
                        Edge(source="CustomCodeGenerator", target="AnswerFormatter"),
                    ],
                    entry_node_ids=["Input"],
                    final_node_id="AnswerFormatter",
                    description="Minimal MBPP round_1 workflow.",
                )
        \end{lstlisting}
    \end{tcolorbox}

\subsection{Minimal initial workflow graph on HotpotQA dataset}
    \begin{tcolorbox}[rounded corners, breakable, colframe=black!70, colback=white, boxrule=4pt, boxsep=0.5pt, enhanced, shadow={3pt}{-3pt}{0pt}{opacity=0.3}]
        \begin{lstlisting}[breaklines=true,showstringspaces=false,language=python]
        from src.core.edge import Edge
        from src.core.graphflow import GraphFlow
        from src.core.nodes.answer_format_node import AnswerFormatNode
        from src.core.nodes.answer_generate_node import AnswerGenerateNode
        from src.core.nodes.input_node import InputNode
        from src.core.workflow import BaseWorkflow
        
        class Workflow(BaseWorkflow):
            def _build_graph(self) -> GraphFlow:
                Input = InputNode(
                    node_id="Input",
                    node_llm_config=self.llm_config,
                    description="Graph input entry.",
                )
                AnswerGenerator = AnswerGenerateNode(
                    node_id="AnswerGenerator",
                    node_llm_config=self.llm_config,
                    description="Built-in answer generation node for direct step-by-step solving.",
                )
                AnswerFormatter = AnswerFormatNode(
                    node_id="AnswerFormatter",
                    dataset_name=self.dataset,
                    node_llm_config=self.llm_config,
                    description="Format the final answer for the target dataset.",
                )
                return GraphFlow(
                    nodes=[Input, AnswerGenerator, AnswerFormatter],
                    edges=[
                        Edge(source="Input", target="AnswerGenerator"),
                        Edge(source="AnswerGenerator", target="AnswerFormatter"),
                    ],
                    entry_node_ids=["Input"],
                    final_node_id="AnswerFormatter",
                    description="Minimal HotpotQA round_1 workflow.",
                )
        \end{lstlisting}
    \end{tcolorbox}

\end{onecolumn}

\end{document}